\documentclass[conference]{IEEEtran}
\IEEEoverridecommandlockouts
\usepackage{times}
\usepackage{soul}
\usepackage{url}
\usepackage[hidelinks]{hyperref}
\usepackage[utf8]{inputenc}
\usepackage[small]{caption}
\usepackage{amsthm}
\usepackage{cite}
\usepackage{amsmath,amssymb,amsfonts}
\usepackage{algorithm}
\usepackage{algorithmic}
\usepackage[switch]{lineno}
\usepackage{graphicx}
\usepackage{textcomp}

\usepackage{threeparttable}
\usepackage{siunitx}
\usepackage{booktabs}
\usepackage{longtable}
\usepackage{graphicx}
\usepackage{tikz}
\usepackage{pifont}   
\usepackage{tablefootnote} 
\usepackage{color}    

\newcommand{\qm}{\textcolor[HTML]{0402e1}{\textbf{\scriptsize\ding{108}}}}
\newcommand{\md}{\textcolor[HTML]{e06e2f}{\textbf{\scriptsize\ding{108}}}}
\newcommand{\oc}{\textcolor[HTML]{a961c8}{\textbf{\scriptsize\ding{108}}}}
\newcommand{\pcqm}{\textcolor[HTML]{55b838}{\textbf{\scriptsize\ding{108}}}}
\newcommand{\mol}{\textcolor[HTML]{3f68a0}{\textbf{\scriptsize\ding{108}}}}

\newcommand{\zinc}{\textcolor[HTML]{337356}{\textbf{\scriptsize\ding{108}}}}

\newcommand{\gdsc}{\textcolor[HTML]{337356}{\textbf{\scriptsize\ding{110}}}}
\newcommand{\ccle}{\textcolor[HTML]{FF5733}{\textbf{\scriptsize\ding{110}}}}
\newcommand{\tcga}{\textcolor[HTML]{8E44AD}{\textbf{\scriptsize\ding{110}}}}
\newcommand{\prism}{\textcolor[HTML]{3498DB}{\textbf{\scriptsize\ding{110}}}}
\newcommand{\nci}{\textcolor[HTML]{E74C3C}{\textbf{\scriptsize\ding{110}}}}
\newcommand{\cosmic}{\textcolor[HTML]{16A085}{\textbf{\scriptsize\ding{110}}}}
\newcommand{\pubchem}{\textcolor[HTML]{F1C40F}{\textbf{\scriptsize\ding{110}}}}

\newcommand{\ctrp}{\textcolor[HTML]{2ECC71}{\textbf{\scriptsize\ding{110}}}}
\newcommand{\pdtc}{\textcolor[HTML]{E67E22}{\textbf{\scriptsize\ding{110}}}}
\newcommand{\scRNA}{\textcolor[HTML]{A61E22}{\textbf{\scriptsize\ding{110}}}}
\newcommand{\depmap}{\textcolor[HTML]{0e00d9}{\textbf{\scriptsize\ding{110}}}}
\newcommand{\OGBbiokg}{\textcolor[HTML]{A58745}{\textbf{\scriptsize\ding{110}}}}
\newcommand{\ZhangDDI}{\textcolor[HTML]{F78045}{\textbf{\scriptsize\ding{110}}}}
\newcommand{\PDBbind}{\textcolor[HTML]{C78045}{\textbf{\scriptsize\ding{110}}}}
\newcommand{\Metz}{\textcolor[HTML]{D21015}{\textbf{\scriptsize\ding{110}}}}
\newcommand{\LIT}{\textcolor[HTML]{E12015}{\textbf{\scriptsize\ding{110}}}}
\newcommand{\Karimi}{\textcolor[HTML]{F29115}{\textbf{\scriptsize\ding{110}}}}

\newcommand{\SB}{\textcolor[HTML]{1402e1}{\textbf{\scriptsize\ding{108}}}}
\newcommand{\UD}{\textcolor[HTML]{5a6e2f}{\textbf{\scriptsize\ding{108}}}}
\newcommand{\UP}{\textcolor[HTML]{9061c8}{\textbf{\scriptsize\ding{108}}}}
\newcommand{\UB}{\textcolor[HTML]{ff02e1}{\textbf{\scriptsize\ding{108}}}}

\newcommand{\MultiClass}{\textcolor[HTML]{CF1651}{\textbf{\scriptsize\ding{73}}}}

\newcommand{\Regression}{\textcolor[HTML]{9B59B6}{\textbf{\scriptsize\ding{72}}}}
\newcommand{\Classification}{\textcolor[HTML]{2ECC71}{\textbf{\scriptsize\ding{73}}}}
\newcommand{\Recommendation}{\textcolor[HTML]{E67E22}{\textbf{\scriptsize\ding{74}}}}

\def\BibTeX{{\rm B\kern-.05em{\sc i\kern-.025em b}\kern-.08em
    T\kern-.1667em\lower.7ex\hbox{E}\kern-.125emX}}
\begin{document}

\title{Graph-structured Small Molecule Drug Discovery Through Deep Learning: Progress, Challenges, and Opportunities}


\author{\IEEEauthorblockN{Kun Li$^{1}$\IEEEauthorrefmark{1}, 
Yida Xiong$^{1}$\IEEEauthorrefmark{1}, 
Hongzhi Zhang$^{1}$\IEEEauthorrefmark{1}, 
Xiantao Cai$^{1}$, 
Jia Wu$^{2}$,
Bo Du$^{1}$, 
Wenbin Hu$^{1}$\IEEEauthorrefmark{2}\thanks{\IEEEauthorrefmark{2}Corresponding author}}
\IEEEauthorblockA{$^{1}$\textit{School of Computer Science, Wuhan University}, Wuhan, China \\
$^{2}$\textit{Department of Computing, Macquarie University}, Sydney, Australia \\
$^{1}$\{likun98, yidaxiong, zhanghongzhi, caixiantao, dubo, hwb\}@whu.edu.cn, $^{2}$jia.wu@mq.edu.au}
\thanks{\IEEEauthorrefmark{1}These authors contributed to the work equally and should be regarded as co-first authors}
}


\maketitle

\begin{abstract}
Due to their excellent drug-like and pharmacokinetic properties, small molecule drugs are widely used to treat various diseases, making them a critical component of drug discovery.  In recent years, with the rapid development of deep learning (DL) techniques, DL-based small molecule drug discovery methods have achieved excellent performance in prediction accuracy, speed, and complex molecular relationship modeling compared to traditional machine learning approaches. These advancements enhance drug screening efficiency and optimization and provide more precise and effective solutions for various drug discovery tasks. Contributing to this field's development, this paper aims to systematically summarize and generalize the recent key tasks and representative techniques in graph-structured small molecule drug discovery in recent years. Specifically, we provide an overview of the major tasks in small molecule drug discovery and their interrelationships. Next, we analyze the six core tasks, summarizing the related methods, commonly used datasets, and technological development trends. Finally, we discuss key challenges, such as interpretability and out-of-distribution generalization, and offer our insights into future research directions for small molecule drug discovery.
\end{abstract}



\section{Introduction}

\begin{figure*}[h]
    \centering
    \includegraphics[width=0.9\linewidth]{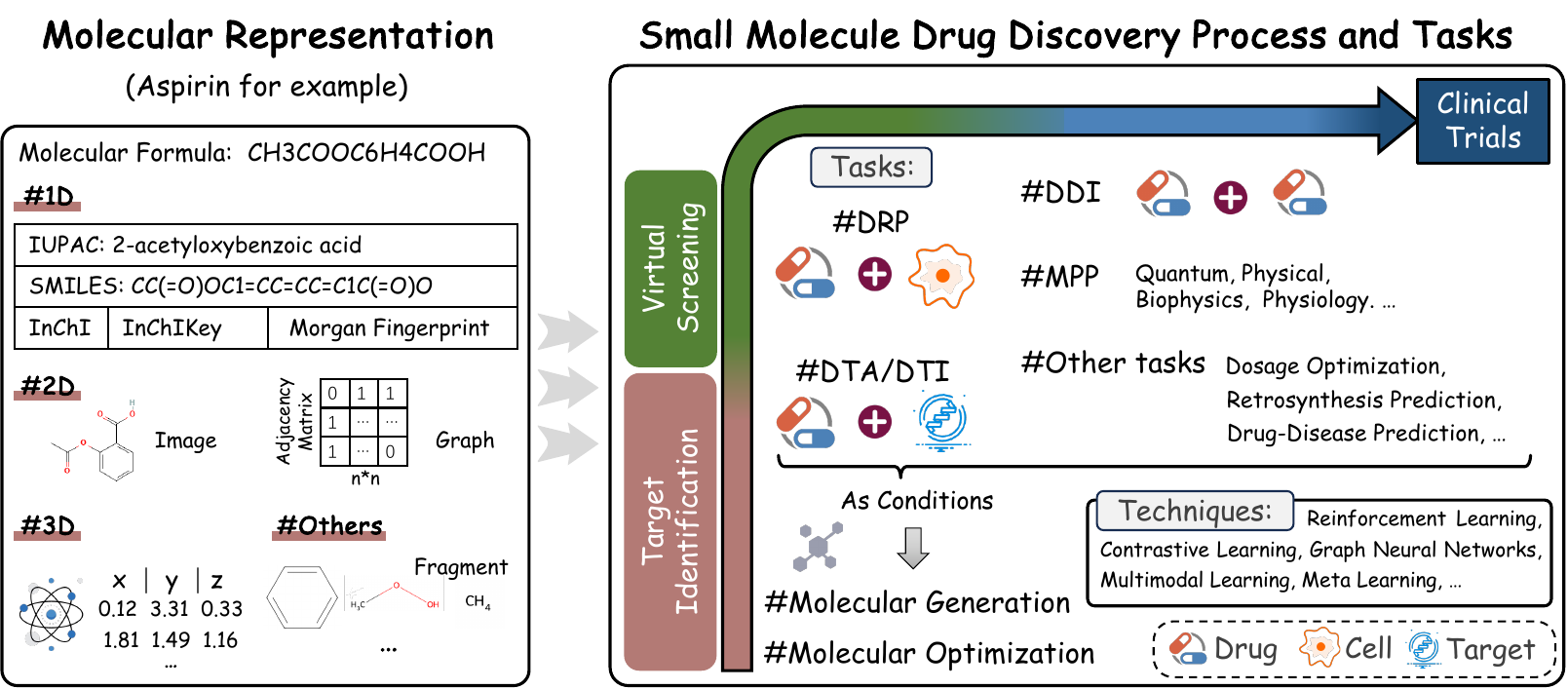}
    \caption{Molecular representation methods and their drug discovery applications.}
    \label{fig:1}
\end{figure*}

Small molecule drugs are chemically synthesized organic compounds with a molecular weight of less than 1,000. Due to their favorable drug-like and pharmacokinetic properties, these drugs play a critical role in the treatment of various diseases, accounting for about 98\% of the total number of commonly used drugs, and are the foundation of modern drug discovery. Small molecule drug discovery encompasses lead compound screening, optimization, biochemical property prediction, and so on. 

\begin{table*}[ht!]
    \centering
    \begin{threeparttable}
    
    \resizebox{0.95\linewidth}{!}{ 
    \begin{tabular}{ccccccrr} 
    \toprule
    \textbf{Methods} &  \# \textbf{Drug}  &  \# \textbf{DrugEn.}&  \# \textbf{Target} &  \# \textbf{TargetEn.} & \# \textbf{FusionM.} & \# \textbf{Data.\&Tasks} & \# \textbf{OOD} \\ \midrule
    DrugCLIP$^{\text{\cite{gao2024drugclip}}}$  &3D  & Uni-mol\cite{Uni-Mol}  & 3D Pocket & Uni-mol  & -  & \cosmic \LIT: \Classification  &  \UP \SB \\
    APLM $^{\text{\cite{sledzieski2022adapting}}}$ &MFP & MLP  & Seq.   & PLM  & Concat  & \gdsc \pdtc \scRNA: \Classification  &  \UD \UP \SB \\
    DTI-MGNN$^{\text{\cite{li2022drug}}}$ &SMILES & KGE  & Seq.   & KGE  & Attention  & \tcga: \Classification  & \SB \\
    HyperAttentionDTI$^{\text{\cite{zhao2022hyperattentiondti}}}$ &SMILES & CNNs  & Seq.   & CNNs  & Attention  &  \gdsc \ccle \tcga: \Classification  & \SB \\
    Drug-VQA$^{\text{\cite{zheng2020predicting}}}$ &SMILES & LSTM  & Distance Map   & CNNs  & Concat  & \cosmic \pubchem  \ctrp: \Regression  & \SB   \\
    MolTrans$^{\text{\cite{huang2021moltrans}}}$  &SMILES & Trans.  & Seq.   & Trans.  & MP  & \gdsc \pdtc \scRNA: \Classification   & \UD \UP   \\
    SiamDTI$^{\text{\cite{zhang2024cross}}}$ &SMILES & Trans.  & Seq.   & Trans.  & BAN  &  \pdtc\scRNA: \Classification   & \UB \UD \UP \SB   \\
    DTIAM $^{\text{\cite{lu2025dtiam}}}$ &Graph & Trans.  & Seq.   & Trans.  & Concat  & \gdsc \ccle : \Classification \Regression   & \UD \UP   \\
    DrugBAN$^{\text{\cite{bai2023interpretable}}}$ &Graph & GNNs  & Seq.   & CNNs  & BAN  & \ctrp \pdtc: \Classification  &  \UB \SB \\
    PSC-CPI$^{\text{\cite{wu2024psc}}}$  &Graph & GNNs  &  Seq., Graph & HRNN, GAT  & MP & \gdsc \pdtc \Metz \Karimi: \Regression  & \UB \UD \UP \SB \\
    Cross-interaction$^{\text{\cite{you2022cross}}}$ &Graph & GNNs  & Seq., Graph   & HRNN, GAT  & Concat  & \pdtc \PDBbind: \Classification  & \UB \UD \UP \SB   \\
    MGNDTI$^{\text{\cite{peng2024mgndti}}}$ &SMILES, Graph & GCNs, RN  &  Seq.& RN  & GN & \pubchem \ctrp \pdtc : \Classification  & \UB \UD \UP \SB \\
    Perceiver-CPI$^{\text{\cite{nguyen2023perceiver}}}$ &SMILES, MFP & GNNs, MLP  & Seq.   & CNN  & Attention  & \gdsc \ccle \PDBbind \Metz: \Regression  & \UB \UD \UP \SB \\

    \bottomrule
    \end{tabular}
    }
    \end{threeparttable} 
    
    \begin{minipage}{0.94\linewidth}
    \scriptsize
     
     \vspace{3pt}
     \textbf{Notations}: 
     
     \textcircled{\raisebox{-0.9pt}{1}} Drug: morgan fingerprint (\textbf{MFP}), 3D conformation (\textbf{3D}).
     
     \textcircled{\raisebox{-0.9pt}{2}} Techiques: Retentive Networks (\textbf{RN}), Directed Message-Passing Neural Network (\textbf{DMPNN}); Protein Language Model (\textbf{PLM}), Hierarchical Recurrent Neural Networks (\textbf{HRNN}); Manhattan Product (\textbf{MP}), Bilinear Attention Network (\textbf{BAN}),  Gating Network (\textbf{GN}), Neural Factorization Machine (\textbf{NFM}), Transformer (\textbf{Trans.}).
    
    \textcircled{\raisebox{-0.9pt}{3}} Datasets: \gdsc~Davis, \ccle~KIBA,  \tcga~DrugBank, \cosmic~DUDE , \pubchem~C.elegans, \ctrp~Human, \pdtc~BindingDB, \scRNA~BioSNAP, \Metz~Metz, \LIT~LIT-PCBA, \Karimi~Karimi.
    
    \textcircled{\raisebox{-0.9pt}{4}} Tasks: \Classification~classification task, \Regression~regression task.
    
    \textcircled{\raisebox{-0.9pt}{5}} Out of distribution: \SB~\textbf{Seen-Both}, \UD~\textbf{Unseen-Drug}, \UP~\textbf{Unseen-Target}, \UB~\textbf{Unseen-Both}.

    \end{minipage} 
    \caption{Summary of representative DTI/DTA tasks methods.}
    \label{tab:DTI}

\end{table*}

It also comprises several key stages, as shown in Figure \ref{fig:1}, including drug–target interaction and affinity (DTI/DTA), drug–cell response (DRP), drug–drug interaction (DDI), molecular property prediction (MPP). They also include molecular generation  (MG) and optimization (MO). These tasks primarily focus on three objects: small molecule drugs, cell lines with omics data, and target proteins. However, the representation methods for these objects are specific and vary considerably, customized approaches. Specifically, the DTI/DTA (target-based drug discovery) and DRP (phenotypic-based drug discovery) tasks are regarded as the two primary lead compound discovery approaches. Accordingly, the MPP and DDI tasks are primarily used to evaluate the physicochemical properties, pharmacokinetics, and potential adverse drug–drug interactions of small molecules. Subsequently, the MG and MO tasks are primarily aimed at expanding the drug candidate pool. In contrast, condition-based generation and optimization, leverage the predictive capabilities of the four tasks—DTI/DTA, DRP, MPP, and DDI—to design molecules with specific properties. These six tasks are closely interconnected and form the core of small molecule drug discovery. 

In recent years, the rapid development of deep learning (DL) technology has considerably improved small molecule drug discovery as computational power increases and data accumulates. DL-based methods have significantly surpassed traditional machine learning techniques in improving prediction accuracy, accelerating computation, enhancing molecular generation and optimization, and modeling complex molecular relationships. For example, deep (DNNs), convolutional (CNNs), and graph neural networks (GNNs) have been widely applied in automated feature extraction and multi-task learning, enabling the identification of potential patterns and improving prediction accuracy in large-scale biomedical data. These advances have improved drug screening efficiency and provided more accurate and effective solutions for various drug discovery tasks. Therefore, the rapid development of DL technology combined with increasingly rich small molecule datasets advances the drug discovery process.


Graph-based representations offer inherent advantages for small molecules, and related studies have introduced numerous methods and datasets for small molecule drug discovery. Early surveys have struggled to meet the needs of current research. To fill this gap, this paper systematically summarizes the key tasks and representative techniques in DL-based small molecule drug discovery over the past three years. According to the stages and requirements of small molecule drug discovery, we provide a detailed overview of various methods for six major tasks—DTI/DTA, DRP, DDI, MPP, MG, and MO—along with relevant datasets and specific technological trends to each task in Section \ref{sec:3}. Given the dataset complexity and diversity in drug discovery, this paper also summarizes the main small molecule datasets with access URLs (see Section \ref{sec:4} for details). Furthermore, we analyze the challenges associated with these tasks and provide insights into potential future research directions in Section \ref{sec:5}. To the best of our knowledge, this paper is the first attempt to present a systematic and comprehensive review of recent DL advancements in small molecule drug discovery.

\begin{table*}[htpb]
\centering
\renewcommand\arraystretch{1.2} 
\setlength\tabcolsep{5pt} 
\begin{threeparttable}
\resizebox{0.98\textwidth}{!}{
\begin{tabular}{ccccccr}
\toprule

\textbf{Methods} & \# \textbf{Cell profiles} & \# \textbf{Drug} & \# \textbf{Cell Line Arch.} & \# \textbf{Drug Arch.} & \# \textbf{Tech.} & \textbf{Datasets\&Tasks} \\

\midrule

DeepTTA\tnote{\cite{deeptta}} & E & Subcomponent & MLP & Trans. & Supervised Learning & \gdsc: \Classification \\
SubCDR\tnote{\cite{subcdr}} & E, R & Subcomponent & CNNs & CNNs & Supervised Learning & \gdsc\cosmic\pubchem: \Classification\Regression \\
TGSA\tnote{\cite{TGSA}} & M, E, C, Net & Graph & GAT & GIN & Supervised Learning & \gdsc: \Regression \\ 
DIPK\tnote{\cite{DIPK}} & E, Net & Graph & GAE & GNNs, Trans. & Pre-training & \gdsc\ccle: \Regression \\
GraphCDR\tnote{\cite{graphcdr}} & M, E, C & Graph & DNNs & GNNs & Contrastive Learning & \gdsc\ccle: \Classification \\
CLDR\tnote{\cite{cldr}} & No constr. & No constr. & Trans. & Trans. & Contrastive Learning & \gdsc: \Regression \\
MSDA\tnote{\cite{MSDA}} & No constr. & No constr. & No constr. & No constr. & Domain Generalization  & \gdsc\nci: \Regression \\
CODE-AE\tnote{\cite{code-ae}} & R & Not req. & AE & Not req. & Domain Generalization & \ccle\tcga\pdtc: \Classification \\
scDEAL\tnote{\cite{scDEAL}} & R, S & Not req. & AE & Not req. & Transfer Learning & \gdsc\ccle\scRNA: \Classification \\
WISER\tnote{\cite{WISER}} & R & Not req. & MLP & Not req. & Weak Supervision Learning & \tcga\depmap: \Classification \\ \bottomrule

\end{tabular}
}
\end{threeparttable}

\begin{minipage}{0.97\linewidth}
\scriptsize 
\vspace{3pt}
\textbf{Notations:}

\textcircled{\raisebox{-0.9pt}{1}} Datasets: \gdsc~GDSCv1/2, \ccle~CCLE, \tcga~TCGA,  \nci~NCI-60, \cosmic~COSMIC, \pubchem~PubChem, \pdtc~PDTC, \depmap~DepMap.

\textcircled{\raisebox{-0.9pt}{2}} Profiles: mutation status (\textbf{M}), gene expression profiles (\textbf{E}), copy number variation (\textbf{C}), RNA sequence (\textbf{R}), single-cell RNA sequence (\textbf{S}), gene interaction network/protein–protein association network from STRING dataset (\textbf{Net}), no need for special module or data (\textbf{Not req.}),  No restrictions on the type of module or data (\textbf{no constr.}).

\textcircled{\raisebox{-0.9pt}{3}} Techiques: Autoencoder (\textbf{AE}), Graph Autoencoder (\textbf{GAE}).
\end{minipage} 

\caption{Summary of representative DRP task methods.}
\label{DRP_table}
\end{table*}

\section{Problem Formulation}


A molecule $M$ with $N$ nodes can be represented by $\mathcal{G}=(\mathcal{X},\mathcal{A})$, where $\mathcal{X} \in \mathbb{R}^{N \times F}$ denotes node features and $\mathcal{A} \in \mathbb{R}^{N \times N}$ signifies the adjacency matrix ($F$ is the node feature dimension). Three-dimensional (3D) molecular structures are commonly represented as 3D graphs $\mathcal{G}_{3D}=(\mathcal{X},\mathcal{A},\mathcal{R})$ or point clouds $\mathcal{P}_{3D}=(\mathcal{X},\mathcal{R})$, where information on the node position in a coordinate system ($\mathbf{r}_{i} \in \mathcal{R}$) is encoded ($\mathcal{R}$ refers to the set of coordinates). 

In addition, the six main tasks in small molecule drug discovery can be abstractly represented as follows: 

The MPP task predicts the properties of a molecule $M$, such as solubility, polarity, and toxicity. This can be represented as:
\begin{equation}
    \text{Y}_{\text{MPP}} = f_{\text{model}}(M),
\end{equation}
\noindent where $f_{\text{model}}(\cdot)$ indicates the respective task's prediction model. The DRP, DTI/DTA, and DDI tasks can be collectively referred to as the \textbf{drug-X} format. Specifically, this format predicts whether two entities, $M$ and $X$ (where $X$ can be another drug $M$, a target $T$, or a cell line $C$) interact. It is can be formulated as:  
\begin{equation}
    Y_{\text{DX}} = f_{\text{model}}(M, X),
\end{equation}
Finally, the MG task generates new molecules $M_{\text{Gen}}$ with/without target property $C$, while the MO task is to optimise the existing molecule $M$ to $M_{\text{Opt}}$, improving $M$'s properties $C$ while preserving its activity. Both tasks can be unified and represented as:
\begin{equation}
    M_{\text{Gen}/\text{Opt}} = \arg \min_{M} f_{\text{val}}(\emptyset /M, \emptyset / C).
\end{equation}
\noindent where $f_{\text{val}}(\cdot)$ is the molecular evaluation function, and $\emptyset$ represents empty set.

\section{Methods}
\label{sec:3}

\subsection{Drug–Target Interaction and Affinity Prediction}
\label{3.1}

The DTI and DTA prediction tasks are extremely crucial for drug discovery. Typically, DTI prediction is formulated as a binary classification problem that determines a drug's interaction with a specific target, outputting a binary label. In contrast, DTA prediction is a regression problem focusing on predicting the binding affinity between a drug and a target, often quantified by metrics such as the dissociation constant (Kd) value. The input for both tasks consists of drug data (e.g., simplified molecular input line entry specification (SMILES), molecular fingerprints, etc.) and target data (e.g., protein sequences, three-dimensional (3D) structure, etc.).

Moreover, DTI/DTA frameworks employ dual-encoder architectures for molecular and protein feature extraction, coupled with interaction or affinity predictors. Table \ref{tab:DTI} displays various DTI/DTA methods based on model structure differences. Early methods, such as CNNs, were inadequate in capturing the structural details of molecules. Meanwhile, GNNs have become the dominant paradigm for processing molecular data due to their ability to characterize the two-dimensional structure of molecules. Knowledge graph-based methods have improved performance by integrating multi-entity relationships (e.g., drugs, targets, and diseases). 

In recent years, out-of-distribution (OOD) data has emerged as a primary research focus. During the DTI/DTA task, test scenarios can be categorized into four types based on whether the drugs and targets in the test set are observed during training: \textbf{Seen-Both}: both the drugs and targets in the test set are present during training. \textbf{Unseen-Drug}: the drugs in the test set are not present in the training one. \textbf{Unseen-Target}: the targets in the test set are not present during training. \textbf{Unseen-Both}: both the drugs and targets in the test set are not present in the training one. All cases except Seen-Both represent OOD problems. For a detailed discussion of OOD issues, refer to Section \ref{5.2}. The predominant approaches to addressing this issue involve domain adaptation and pre-training techniques. Public databases, including Davis and KIBA offer an extensive array of binding data (for more datasets details, see Table \ref{Datasets}).

\subsection{Drug–Cell Response Prediction}
\label{3.2}




\begin{table*}[ht!]
\centering
\renewcommand\arraystretch{1.1} 
\setlength\tabcolsep{12pt} 
\begin{threeparttable}
\resizebox{\textwidth}{!}{
\begin{tabular}{cccl|cccl}
\toprule
\textbf{Methods} & \# \textbf{Rep.} & \# \textbf{Arch.} & \textbf{Pre-train(Data.)} & \textbf{Methods} & \# \textbf{Rep.} & \# \textbf{Arch.} & \textbf{Pre-train(Data.)} \\ \midrule
GEM\tnote{\cite{GEM}} & Graph, 3D & GNNs & SSL, \textcolor[HTML]{9a2e61}{\textbf{\scriptsize\ding{115}$^\textbf{1-9}$}}\textcolor[HTML]{3069a5}{\textbf{\scriptsize\ding{115}$^\textbf{1-6}$}} & SliDe\tnote{\cite{SliDe}} &  3D & TorchMD-Net & DN, (\pcqm) \qm\md \\
GraphMVP\tnote{\cite{GraphMVP}} & Graph, 3D & GNNs & SSL, (\pcqm) \qm\md\textcolor[HTML]{9a2e61}{\textbf{\scriptsize\ding{115}$^\textbf{1-8}$}} &  Uni-Mol\tnote{\cite{Uni-Mol}} & 3D & Trans. & DN, \textcolor[HTML]{9a2e61}{\textbf{\scriptsize\ding{115}$^\textbf{1-9}$}} \textcolor[HTML]{3069a5}{\textbf{\scriptsize\ding{115}$^\textbf{1-6}$}} \\
Transformer-M\tnote{\cite{Transformer-M}} & Graph, 3D & Trans. & DN, (\pcqm) \qm & MolE\tnote{\cite{MolE}} & Graph & Trans. & SL, TDC \\
MolSpectra\tnote{\cite{MolSpectra}} & Graph, 3D & Trans. & DN, SSL, (\pcqm\textcolor[HTML]{3069a5}{\textbf{\scriptsize\ding{116}}}) \qm\md & MolMCL\tnote{\cite{MolMCL}} & Graph & GNNs & SSL, \textcolor[HTML]{9a2e61}{\textbf{\scriptsize\ding{115}$^\textbf{1-7}$}} \\
Uni-Mol+\tnote{\cite{Uni-Mol+}} & Graph, 3D & Trans. & SL, \pcqm\oc & MolCLR\tnote{\cite{MolCLR}} & Graph & GNNs & CL, \textcolor[HTML]{9a2e61}{\textbf{\scriptsize\ding{115}$^\textbf{1-7}$}}\textcolor[HTML]{3069a5}{\textbf{\scriptsize\ding{115}$^\textbf{1-6}$}} \\
MOLEBLEND\tnote{\cite{MOLEBLEND}} & Graph, 3D & Trans. & SL, \pcqm & TopExpert\tnote{\cite{TopExpert}} & Graph & GNNs & SSL, (\pcqm) \textcolor[HTML]{9a2e61}{\textbf{\scriptsize\ding{115}$^\textbf{1-8}$}}  \\
Uni-Mol2\tnote{\cite{Uni-Mol2}} & Graph, 3D & Trans. & SSL, \qm\textcolor[HTML]{833c5f}{\textbf{\scriptsize\ding{108}}}& GraphMAE\tnote{\cite{GraphMAE}} & Graph & GNNs & SSL, \textcolor[HTML]{9a2e61}{\textbf{\scriptsize\ding{115}$^\textbf{1-6}$}}\\
TGT\tnote{\cite{TGT}} & Graph & GT & DN, SSL, \qm\oc & InstructMol\tnote{\cite{InstructMol}} & Graph & GNNs & Semi-SL, \textcolor[HTML]{9a2e61}{\textbf{\scriptsize\ding{115}$^\textbf{1-6}$}}\textcolor[HTML]{3069a5}{\textbf{\scriptsize\ding{115}$^\textbf{1-6}$}} \\

\bottomrule

\end{tabular}}
\end{threeparttable}

\begin{minipage}{0.99\linewidth}
\scriptsize 
 \vspace{3pt}
 \textbf{Notations:}
 
 \textcircled{\raisebox{-0.9pt}{1}} Datasets:  \qm~QM9; \md~MD17/22; \pcqm~PCQM4Mv2;  \textcolor[HTML]{833c5f}{\textbf{\scriptsize\ding{108}}}~COMPAS-1D; \textcolor[HTML]{3069a5}{\textbf{\scriptsize\ding{116}}}~QM9Spectra; \textcolor[HTML]{3069a5}{\textbf{\scriptsize\ding{115}$^\textbf{6}$}}~Regression datasets, including FreeSolv, ESOL, Lipo, QM7, QM8, and QM9; \textcolor[HTML]{9a2e61}{\textbf{\scriptsize\ding{115}$^\textbf{9}$}}~Classification datasets, including BBBP, Tox21, ClinTox, HIV, BACE, SIDER, MUV, ToxCast, and PCBA. TDC is \href{https://tdcommons.ai}{therapeutics data commons platform}.
 
 \textcircled{\raisebox{-0.9pt}{2}} Methods: Self-supervised Learning (\textbf{SSL}), Denoising Pre-training (\textbf{DN}), Graph Transformer (\textbf{GT}), Semi-Supervised Learning (\textbf{Semi-SL}).

\end{minipage} 

\caption{Summary of representative MPP task methods with pre-training.}
\label{MPP_W_PRE}
\end{table*}


\begin{table*}[ht!]
\centering
\renewcommand\arraystretch{1.1} 
\setlength\tabcolsep{12pt} 
\begin{threeparttable}
\resizebox{\textwidth}{!}{
\begin{tabular}{cccl|cccl}
\toprule
\textbf{Methods} & \# \textbf{Rep.} & \# \textbf{Arch.} & \textbf{Dataset} & \textbf{Model Name} & \# \textbf{Rep.} & \# \textbf{Arch.} & \textbf{Dataset} \\ \midrule
SphereNet\tnote{\cite{SphereNet}} & 3D graph & GNNs & \qm\md\oc  & SR-GINE\tnote{\cite{srgine}} & 1D Desc & GIN & \textcolor[HTML]{9a2e61}{\textbf{\scriptsize\ding{115}$^\textbf{1-6}$}}\textcolor[HTML]{3069a5}{\textbf{\scriptsize\ding{115}$^\textbf{1-5}$}}\textcolor[HTML]{d1743f}{\textbf{\scriptsize\ding{116}$^\textbf{6}$}}  \\
ComENet\tnote{\cite{ComENet}} & 3D graph & GNNs & \qm\oc\mol & FragNet\tnote{\cite{fragnet}} & Frag graph & GNNs & \qm\zinc\textcolor[HTML]{6fb64c}{\textbf{\scriptsize\ding{116}$^\textbf{1-2}$}}  \\
TorchMD-Net\tnote{\cite{TorchMD-NET}} & 3D graph & GNNs & \qm\md\textcolor[HTML]{9f64c2}{\textbf{\scriptsize\ding{115}}} & Polymer Walk\tnote{\cite{walk}} & Motif graph & GNNs & \textcolor[HTML]{d1743f}{\textbf{\scriptsize\ding{116}$^\textbf{1-2}$}} \\
SCHull\tnote{\cite{schull}} & 3D graph & GNNs & \md &  Geo-DEG\tnote{\cite{Geo-DEG}} & Graph & GNNs & \textcolor[HTML]{3069a5}{\textbf{\scriptsize\ding{115}$^\textbf{1,3}$}}\textcolor[HTML]{d1743f}{\textbf{\scriptsize\ding{116}$^\textbf{1-5}$}}  \\
Equiformer\tnote{\cite{Equiformer}} & 3D graph & GT & \qm\md\oc & GS-Meta\tnote{\cite{gsmeta}} & Graph & GNNs & \textcolor[HTML]{9a2e61}{\textbf{\scriptsize\ding{115}$^\textbf{2-3,6-9}$}} \\
ViSNet\tnote{\cite{ViSNet}} & Graph, 3D & Trans. & \qm\mol  & GraphSAM\tnote{\cite{GraphSAM}} & Graph & GT & \textcolor[HTML]{9a2e61}{\textbf{\scriptsize\ding{115}$^\textbf{1-3,6}$}}\textcolor[HTML]{3069a5}{\textbf{\scriptsize\ding{115}$^\textbf{2-3}$}} \\

\bottomrule

\end{tabular}
}
\end{threeparttable}

\begin{minipage}{0.99\linewidth} \scriptsize
\vspace{3pt}
 \textbf{Notations:}
 \textcircled{\raisebox{-0.9pt}{1}} Datasets:  \oc~OC20, \mol~Molecule3D, \textcolor[HTML]{d1743f}{\textbf{\scriptsize\ding{115}}}~3BPA, \textcolor[HTML]{9f64c2}{\textbf{\scriptsize\ding{115}}}~ANI-1. Datasets for materials science: HOPV, PTC, CROW, Permeability, DILI, and OCELOT Chromophore  (\textcolor[HTML]{d1743f}{\textbf{\scriptsize\ding{116}$^\textbf{6}$}}). Peptides-struct/-func and LogP (\textcolor[HTML]{6fb64c}{\textbf{\scriptsize\ding{116}$^\textbf{2}$}}).

 
\end{minipage} 
\caption{Summary of representative MPP task methods without pre-training.}
\label{MPP_WO_PRE}
\end{table*}


The DRP task predicts the response of specific cells to drugs, playing a crucial role in advancing personalized medicine. This task emphasizes the cellular context, making it particularly relevant for precision oncology and targeted therapies. It primarily determines a drug's efficacy against a given cell line, often represented by metrics such as IC50 values.


Recently, DRP methods have made significant breakthroughs by combining techniques such as domain adaptation, multi-modal data integration, and transfer learning. Table \ref{DRP_table} provides a comprehensive summary of these methods' key features and properties. According to the problem definition, DRP can be abstracted as the single property prediction task of a drug molecule under various conditions (i.e., different cell lines). Accordingly, several methods focusing on OOD generalization have been proposed, particularly when treating drugs and cell lines as independent input objects. In the present time, DRP research has increasingly focused on improving model performance for unknown cell lines and drug molecules. This shift is crucial for practical applications in drug screening (i.e., predicting the efficacy of drug molecules not present in the training set) and precision medicine (i.e., predicting the effects of cell line genomics not included in the training set). Therefore, these independent inputs can be viewed as distinct domains. Therefore,  methods based on contrastive and transfer learning have been proposed and widely applied to enhance OOD generalization performance.

Currently, the main DRP datasets include GDSCv1/2 and CCLE (See Table \ref{Datasets} for details). These datasets can be divided into two categories: those primarily containing quantitative drug–cell response values and those focusing on genomic cell data, such as transcriptomics, mutational genomics, and ribonucleic acid (RNA) sequences. This distinction arises because cells in the DRP task can be described using various omics data. The first dataset category emphasizes statistical response values and only reports cell line types. Meanwhile, the second category mainly comprises the genomics data of specific cell lines. Similar to the DTI/DTA task discussed in Section \ref{3.1}, DRP also faces significant challenges related to OOD generalization, as models must accurately predict responses for unseen drug–cell combinations.


\subsection{Drug–Drug Interaction Prediction}
\label{3.3}

\begin{table}[htpb]
    \centering
    \begin{threeparttable}
    \renewcommand\arraystretch{1.0} 
    \resizebox{\linewidth}{!}{ 
    \begin{tabular}{cccr} 
    \toprule

     \textbf{Model}& \# \textbf{Drug} & \# \textbf{Tech.}  & \# \textbf{Data.\&Tasks} \\ \midrule

    ZeroDDI$^{\text{\cite{wang2024zeroddi}}}$   & SMILES  & SEL      &  \gdsc: \Recommendation \\
     MKG-FENN$^{\text{\cite{wu2024mkg}}}$ & SMILES      & KG   & \gdsc: \MultiClass\\
     BI-GNN$^{\text{\cite{bai2020bi}}}$& Graph  & GNN     &\gdsc \prism: \MultiClass  \\

    CGIB$^{\text{\cite{lee2023conditional}}}$  & Graph       & GIB    & \ZhangDDI \pdtc: \MultiClass \Regression \\
    IGIB-ISE$^{\text{\cite{lee2023conditional}}}$  & Graph       & IGIB   & \ZhangDDI \pdtc: \MultiClass \Regression \\
    PHGL-DDI$^{\text{\cite{yuan2025phgl}}}$  & Graph       & CL  & \gdsc \pdtc: \MultiClass  \\
    DANN-DDI $^{\text{\cite{liu2022enhancing}}}$  & Graph       & Attention  & \gdsc \ccle: \MultiClass \\
    
    TIGER$^{\text{\cite{su2024dual}}}$  & Graph  & GNN    & \gdsc \prism  \OGBbiokg: \MultiClass  \\
    GoGNN$^{\text{\cite{wang2020gognn}}}$  & Graph  & GNN    &   \gdsc \cosmic  \ctrp: \Classification\\
    CSGNN$^{\text{\cite{zhao2021csgnn}}}$   & Graph  & CL    & \pdtc: \Classification \\
    DeepDDI$^{\text{\cite{ryu2018deep}}}$     & SMILES, Text     & SSP    & \gdsc: \MultiClass \\
    DDKG $^{\text{\cite{DDKG}}}$  & SMILES, Graph       & KG  & \ccle \OGBbiokg: \MultiClass \\
    
    \bottomrule
    
    \end{tabular}
    }
    \end{threeparttable} 
    
    \begin{minipage}{0.98\linewidth}
    \scriptsize
     
     \vspace{3pt}
     
     \textbf{Notations}:
     
     \textcircled{\raisebox{-0.9pt}{1}} Techniques:  Enhanced Learning (\textbf{SEL}), Knowledge Graph (\textbf{KG}),  Semantic  Graph Information Bottleneck (\textbf{GIB}), Interactive Graph Information Bottleneck (\textbf{IGIB}), Contrastive Learning (\textbf{CL}), Structural Similarity Profile (\textbf{SSP}).
     
    \textcircled{\raisebox{-0.9pt}{2}} Datasets: \gdsc~DrugBank, \ccle~KEGG,  \cosmic~SIDER,   \pdtc~ChCh-Miner, \ZhangDDI~ZhangDDI, \prism DRUGCOMBO, \OGBbiokg OGB-biokg.
    
    \textcircled{\raisebox{-0.9pt}{3}} Tasks: \MultiClass~multi-class classification task, \Recommendation~recommendation task.

    \end{minipage} 
    \caption{Summary of representative DDI task methods.}
    \label{tab:DDI}

\end{table}

The DDI prediction task is critical for identifying adverse pharmacological effects caused by combined drug use. This task mainly involves classifying interaction types between drug pairs, ranging from binary detection (i.e., presence/absence) to multi-class interaction mechanism categorization. 

Table \ref{tab:DDI} provides a comprehensive summary. Notably, GNNs have emerged as the dominant tools for DDI prediction due to their ability to model molecular structures and interactions using graph representations. Knowledge graph methods improve performance by modeling the relationships between drugs, targets, and biological pathways. Recent advancements incorporate self-supervised techniques such as contrastive learning to address data scarcity in OOD scenarios. Grounded in information theory, graph information bottleneck (GIB)\cite{lee2023conditional} methods have demonstrated enhanced prediction accuracy by filtering irrelevant molecular features. Presently these methods highlight the shift toward robust feature learning and explainable models, positioning computational DDI prediction as a scalable complement to traditional approaches. Therefore, continuous progress in multi-modal data (e.g., 3D structures) integration and framework pre-training lead to broader in pharmacological safety assessment applications.  

Public databases can be categorized based on their content and functionality into drug omics, drug adverse effect, and knowledge graph databases. Drug omics databases are primarily composed of drug-related interaction data. Commonly used drug omics databases, such as DrugBank, KEGG, PubChem, and DrugCentral, play a crucial role in DDI prediction. The drug adverse effect datasets, such as TWOSIDES and SIDER are specifically utilized for predicting adverse effects. Finally, DRKG and Bio2RDF represent two comprehensive drug knowledge graph databases contribute significantly contribute to the drug discovery field (See Table \ref{Datasets} for details). The aforementioned datasets contribute to ensuring data feasibility for the DDI tasks. Similarly, DDI also encounters substantial obstacles pertaining to OOD generalization.


\subsection{Molecule Property Prediction}
\label{3.4}

\begin{table*}[htpb]
  \centering
  \small  
  \renewcommand \arraystretch{0.8}
  \begin{tabular}{p{2.3cm}<{\centering}|p{1.2cm}<{\centering}|p{11cm}<{\centering}|p{1.2cm}<{\centering}}
    \toprule
    Models & Category & Backbone & Condition
    \\ \midrule
     $\mathrm{3DLinker}^
     {\text{\cite{huang20223dlinker}}}$ & VAE &
     $ p(\mathcal{G}^{\prime}, \mathcal{M}^{\prime} | \mathcal{G}_{\text{frag}}, \mathcal{M}_{\text{frag}}) $
     & - \\ \midrule
     $\mathrm{GF\mbox{-}VAE}^{\text{\cite{ma2021gf}}}$ & Flow &
     $ p(\mathcal{X}^{\prime} | \mathbf{z}_{\mathcal{X}}, \mathcal{A}) p(A^{\prime} | \mathbf{z}_{\mathcal{A}}) = p(\mathbf{z}_{\mathcal{X}}) \big( \big| \mathrm{det}\big(\frac{\partial f^{atom}_{\omega}(\mathcal{X}^{\prime}, \mathcal{A})}{\partial \mathcal{X}^{\prime}} \big) \big| \big) 
     p(\mathbf{z}_{\mathcal{A}}) \big( \big| \mathrm{det}\big(\frac{\partial f^{bond}_{\theta}(\mathcal{A}^{\prime})}{\partial \mathcal{A}^{\prime}} \big) \big| \big) $
     & Property \\ \midrule
     $\mathrm{MolHF}^{\text{\cite{zhu2023molhf}}}$ & Flow &
     $ \mathcal{G}^{\prime} = (\mathcal{X}^{\prime}, \mathcal{A}^{\prime}) = f^{-1}_{\mathcal{X}|\mathcal{A}}(\mathbf{z}_{\mathcal{X}}, f^{-1}_{\mathcal{A}}(\mathbf{z}_{\mathcal{A}})) $
     & Property \\ \midrule
     $\mathrm{CGCF}^{\text{\cite{xu2021learning}}}$ & Flow & 
     $ p_{\theta}(\mathbf{R} | \mathcal{G}) = \int p(\mathbf{R} | \mathbf{d}, \mathcal{G}) \cdot p_{\theta}(\mathbf{d} | \mathcal{G}) \, d\mathbf{d} $
     & - \\ \midrule
     $\mathrm{LatentGAN}^{\text{\cite{prykhodko2019novo}}}$ & GAN &
     $ p(M^{\prime} | \mathbf{z}_{M}, \mathbf{z}_{fake}) $
     & - \\ \midrule
     $\mathrm{TransORGAN}^{\text{\cite{ijcai2022p539}}}$ & GAN &
     $\begin{cases}
         M^{\prime} = \mathbf{G}_{\theta}(\hat{M}) \\
         \min\limits_{\theta} \max\limits_{\phi} \tilde{V}(\mathbf{G}_{\theta}, \mathbf{D}_{\phi}) = \mathbb{E}_{x\sim p_{\text{data}}(x)}[\log \mathbf{D}_{\phi}(x)] + \mathbb{E}_{z\sim p_{z}(z)}[\log(1-\mathbf{D}_{\phi}(\mathbf{G}_{\theta}(z)))]
     \end{cases}$
     & Property \\ \midrule
     $\mathrm{GDSS}^{\text{\cite{jo2022score}}}$ & Diffusion &
     $\begin{cases}
         d\mathcal{X}_{t} = \mathrm{f}_{1, t}(\mathcal{X}_{t})dt + g_{1, t}d\mathrm{w}_{1} - g_{1, t}^{2}\mathrm{s}_{\theta, t}(\mathcal{X}_{t}, \mathcal{A}_{t})dt \\
         d\mathcal{A}_{t} = \mathrm{f}_{2, t}(\mathcal{A}_{t})dt + g_{2, t}d\mathrm{w}_{2} - g_{2, t}^{2}\mathrm{s}_{\theta, t}(\mathcal{X}_{t}, \mathcal{A}_{t})dt
     \end{cases}$
     & - \\ \midrule
     $\mathrm{GeoLDM}^{\text{\cite{xu2023geometric}}}$ & Diffusion &
     $\begin{cases}
         p_{\theta}(\mathbf{z}_{\mathcal{X}, t-1}, \mathbf{z}_{\mathcal{R}, t-1} | \mathbf{z}_{\mathcal{X}, t}, \mathbf{z}_{\mathcal{R}, t}) = p_{\theta}(\text{C}\mathbf{z}_{\mathcal{X}, t-1}, \mathbf{z}_{\mathcal{R}, t-1} | \text{C}\mathbf{z}_{\mathcal{X}, t}, \mathbf{z}_{\mathcal{R}, t})), \forall \, \text{C} \\
         \text{C}\mathbf{z}_{\mathcal{X}, t-1} + t, \mathbf{z}_{\mathcal{R}, t-1} = \epsilon_{\theta}(\text{C}\mathbf{z}_{\mathcal{X}, t} + t, \mathbf{z}_{\mathcal{R}, t}, t), \forall \, \text{C}, \, t
     \end{cases}$
     & - \\ \midrule     
     $\mathrm{MOOD}^{\text{\cite{lee2023exploring}}}$ & Diffusion &
     $ d \mathcal{G}_{t} = [\mathrm{f}_{t}(\mathcal{G}_{t}) - g_{t}^{2} \nabla_{\mathcal{G}_{t}} \log p_{t}(\mathcal{G}_{t} | \mathbf{y}_{o} = \lambda_{o})] dt + g_{t} dw $
     & Property \\ \midrule
     $\mathrm{DecompDiff}^{\text{\cite{guan2023decompdiff}}}$ & Diffusion &
     $\begin{cases}
         p_{\theta}(\mathbf{x}_{t-1} | \mathbf{x}_{t}, \mathbf{x}_{0}, \mathcal{P}) = \prod_{i=1}^{\mathcal{X}} \sum_{k=1}^{\mathcal{K}} \eta_{ik} \mathcal{N} \left( \mathbf{\tilde{x}}_{t-1,k}^{(i)}; \tilde{\mu}_t \left( \mathbf{\tilde{x}}_{t,k}^{(i)}, \mathbf{\tilde{x}}_{0,k}^{(i)} \right), \tilde{\beta}_{t} \Sigma_k \right) \\
         \tilde{\mu}_t \left( \mathbf{\tilde{x}}_{t,k}^{(i)}, \mathbf{\tilde{x}}_{0,k}^{(i)} \right) = \frac{\sqrt{\alpha_{t}}(1 - \overline{\alpha}_{t-1})}{1 - \overline{\alpha}_{t}} \mathbf{\tilde{x}}_{t,k}^{(i)} + \frac{\sqrt{\overline{\alpha}_{t-1}}\beta_{t}}{1 - \overline{\alpha}_{t}} \mathbf{\tilde{x}}_{0,k}^{(i)}
     \end{cases}$
     & Protein \\ \midrule
     $\mathrm{Graph\,DiT}^{\text{\cite{liu2024graph}}}$ & Diffusion &
     $ \hat{p}_{\theta}(\mathcal{G}_{t-1} | \mathcal{G}, \mathcal{C}) = \log p_{\theta}(\mathcal{G}_{t-1} | \mathcal{G}_{t}) + s(\log p_{\theta}(\mathcal{G}_{t-1} | \mathcal{G}_{t}, \mathcal{C}) - \log p_{\theta}(\mathcal{G}_{t-1} | \mathcal{G}_{t})) $
     & Property 
     \\ \bottomrule
  \end{tabular}
  \begin{minipage}{0.96\linewidth}
    \scriptsize 
    \vspace{3pt}
    \textbf{Notations}: 
    $\mathbf{d}$ denotes a set of pairwise distances between atoms. 
    $\mathcal{G}_{\text{frag}}$ and $\mathcal{M}_{\text{frag}}$ are the graph and geometry per fragment. 
    $f^{atom}_{\omega}$ and $f^{bond}_{\theta}$ are coupling layers which process atoms and bonds respectively. 
    $\mathbf{R}$ denotes the distribution of atomic positions. $p_{\theta}(\mathbf{d} | \mathcal{G})$ models the distribution of inter-atomic distances given the graph $\mathcal{G}$ and $p(\mathbf{R} | \mathbf{d}, \mathcal{G})$ models the distribution of conformations given the distances $\mathbf{d}$. 
    $ \mathbf{G}(\cdot) $ and $ \mathbf{D}(\cdot) $ are the generator and the discriminator, $\hat{M}$ denotes the variant SMILES and $\tilde{V}$ is the value function. 
    $\mathbf{z}_{fake}$ denotes the fake data generated by generator in GAN framework. 
    Meanwhile, $\mathrm{f}_{1, t}(\cdot)$ and $\mathrm{f}_{2, t}(\cdot)$ are linear drift coefficients, $g_{1, t}$ and $g_{2, t}$ are scalar diffusion coefficients, and $\mathrm{w}_{1}$, $\mathrm{w}_{2}$ are reverse-time standard Wiener processes. 
    For a 3D molecule, it can be represented as point clouds $(\mathcal{X}, \mathcal{R})$, where $\mathcal{X}$ is the atom coordinates matrix and $\mathcal{R}$ is the node feature matrix. 
    $\mathbf{y}_{o}$ represents the OOD condition and $\lambda_{o}$ is the parameter controlling the OOD generative process. 
    Finally, $\mathcal{P}$ denotes the set of atoms of the binding site, $\mathcal{K}$ signifies the set of fragments per molecule, and $\Sigma_k \in \mathbb{R}^{3}$ is the prior covariance matrix. $\eta_{ik} = 1$ indicates that the $i$-th molecule atom corresponds to the $k$-th prior. $\alpha_{t}, \beta_{t}, \overline{\alpha}_{t}, \tilde{\beta}_{t}$ are noise parameters. 
    $\mathcal{C}$ represents the generative conditions and $s$ denotes the scale of conditional guidance. 
    $\mu_{\theta}$ is a parameterized mean function consisting of a deblurring network and $\delta$ is the small amount of standard deviation for noise. $\hat{\mathbf{x}}^\mathrm{f}$ denotes the 3D structure of fragment coordinates. 
    
  \end{minipage}
  \caption{Summary of molecular generation methods.}
  \label{tab:MolGen}
\end{table*}

The MPP task predicts a molecule's physical, quantum chemical, and biological properties based on its structure or description. This is essential for applications such as drug design and materials science. Accurately predicting these properties can accelerate the development of new drugs, optimize material properties, and advance scientific research in chemistry and biology.


Various representations, such as SMILES, graph-based representations, and one-dimensional (1D) descriptions (e.g., molecular fingerprints and atom types), are used to encode molecular structures. The representation directly impacts prediction model performance. Tables \ref{MPP_W_PRE} and \ref{MPP_WO_PRE} provide a detailed overview of the various methods with or without pre-training and their technical routes in predicting different molecular properties. Since different methods vary in dataset partitioning, evaluation metrics, random seeds, and implementation details, we selected representative methods and provided detailed descriptions of molecular representation and feature encoding techniques without comparing performance. The recent methods emphasize pre-training techniques (e.g., self-supervised and multi-modal contrastive learning, and 3D-based denoising tasks), which use large datasets to model and optimize the molecular feature space, fine-tuning it according to specific datasets. This pre-training technique can learn generic molecular feature representations from large amounts of labeled (or unlabeled) 3D, textual, or image data, thereby demonstrating excellent generalization performance on small-scale labeled data. In addition, MPP also encounters OOD challenges, especially when dataset splits are based on molecular scaffolds or similarity features, leading to poor predictions for structurally distinct samples.


\begin{table*}[htpb]
  \centering
  \small  
  \renewcommand \arraystretch{0.8}
  \begin{tabular}{p{2.6cm}<{\centering}|p{1.25cm}<{\centering}|p{10cm}<{\centering}|p{1.8cm}<{\centering}}
    \toprule
    Models & Category & Backbone & Condition
    \\ \midrule
     $\mathrm{JTVAE}^{\text{\cite{jin2018junction}}}$ & VAE & 
     $ p(\mathcal{G}^{\prime} | \mathbf{z}_{\mathcal{T}}, \mathbf{z}_{\mathcal{G}}) $
     & Property \\ \midrule
     $\mathrm{FFLOM^{\text{\cite{jin2023fflom}}}}$ & Flow &
     $ \mathcal{G}^{\prime} = f_{\mathcal{G}}^{-1}(\mathbf{z}_{\mathcal{G}}) $
     & Protein \\ \midrule
     $\mathrm{Prompt\mbox{-}MolOpt^{\text{\cite{wu2024leveraging}}}}$ & AutoReg &
     $p(M_{\text{tag}_{t+1}} | M_{\text{tag}_{1:t}}, \mathcal{C}) = \mathrm{Trans}(\text{Emb} + \sum \mathbf{z}_{\mathcal{C}})$
     & Property \\ \midrule
     $\mathrm{Mol\mbox{-}CycleGAN^{\text{\cite{maziarka2020mol}}}}$ & GAN &
     $\begin{cases}
         M^{\prime} = \mathbf{G}_{1}(\mathbf{G}_{2}(M)) \\
         \mathbf{G}_{1}^{*}, \mathbf{G}_{2}^{*} = \mathrm{arg} \min\limits_{\mathbf{G}_{1}, \mathbf{G}_{2}} \max\limits_{\mathbf{D}_{1}, \mathbf{D}_{2}} \tilde{V}(\mathbf{G}_{1}, \mathbf{G}_{2}, \mathbf{D}_{1}, \mathbf{D}_{2})
     \end{cases}$
     & Property \\ \midrule
     $\mathrm{DecompOpt^{\text{\cite{zhoudecompopt}}}}$ & Diffusion &
     $\begin{cases}
         p_{\theta}(M_{0:T-1} | M_{T}, \mathcal{P}, \{ A_{k} \}) = \prod_{t=1}^{T} p_{\theta}(M_{t-1} | M_{t}, \mathcal{P}, \{ A_{k} \}) \\
         M_{0} \sim q(M' | \mathcal{P}, \{ A_{k} \})
     \end{cases}$
     & Protein \\ \midrule
     $\mathrm{PMDM}^{\text{\cite{huang2024dual}}}$ & Diffusion &
     $\begin{cases}
         p_{\theta}(\mathcal{G}_{t-1} | \mathcal{G}_{t}) = \mathcal{N}(\mathcal{G}_{t-1}; \mu_{\theta}(\mathcal{G}_{t}, t), \sigma_{t}^{2}I), p_{\theta}(\mathcal{G}_{0:T-1} | \mathcal{G}_{T}) \\
         \mu_{\theta}(\mathcal{G}_{t}, t) = \frac{1}{\sqrt{1 - \beta_{t}}} \Big ( \mathcal{G}_{t} - \frac{\beta_{t}}{\sqrt{1 - \overline{\alpha}_{t}}} \epsilon_{\theta}(\mathcal{G}_{t}, t) \Big )
     \end{cases}$
     & Protein \\ \midrule

    $\mathrm{FMOP}^{\text{\cite{li2024fragment}}}$ & Diffusion & 
    $\begin{cases}
        \mathcal{A}_{t-1} = W M^{\mathcal{A}} \odot \mathcal{A}_{t-1}^{ukn} + (1-M^{\mathcal{A}}) \odot \mathcal{A}_{t-1}^{kn} \\
        \mathcal{X}_{t-1} = W M^{\mathcal{X}} \odot \mathcal{X}_{t-1}^{ukn} + (1-M^{\mathcal{X}}) \odot \mathcal{X}_{t-1}^{kn} \\
        \tilde{\epsilon}_{\theta}(\mathcal{G}_{t-1}, \mathcal{C}) = \tilde{w} \epsilon_{\theta}(\mathcal{G}_{t} + \epsilon, \mathcal{C}) + (1-\tilde{w})\epsilon_{\theta}(\mathcal{G}_{t} + \epsilon, \emptyset)
    \end{cases}$
    & Cell Line \\ \midrule
     
     $\mathrm{MARS^{\text{\cite{xiemars2021mars}}}}$ & Search &
     $\begin{cases}
         (p_{add}, p_{frag}, p_{del}) = \mathcal{M}_{\theta}(M_{i}^{(t-1)}) \rightarrow \mathcal{S} \\
         M^{\prime} = \mathrm{arg} \max \log M_{\theta}(\mathcal{S})
     \end{cases}$
     & Property \\ \midrule
     $\mathrm{MolEvol^{\text{\cite{chen2021molecule}}}}$ & Search &
     $ p_{\theta}(M) = \int_{s \in \mathcal{S}} p(s) p_{\theta}(M | s) ds $
     & Property \\ \midrule
     $\mathrm{MolSearch^{\text{\cite{sun2022molsearch}}}}$ & Search &
     $\begin{cases}
         \mathbf{U}_{k} = \frac{\mathcal{F}_{k}}{n_{k}} + \lambda \sqrt{\frac{4 \ln n + \ln \mathrm{d}}{2n_{k}}} \quad \mathrm{for} \, k \, \mathrm{=} \, \mathrm{1, ..., K} \\
         V_{p} = \mathrm{P}(\mathbf{U}_{1}, ..., \mathbf{U}_{\mathrm{K}})
     \end{cases}$
     & Property \\ \midrule
     $\mathrm{DST^{\text{\cite{fudifferentiable}}}}$ & Search &
     $ \tilde{\mathcal{X}}_{*}, \tilde{\mathcal{A}}_{*}, \tilde{\mathbf{w}}_{*} = \mathrm{arg} \max_{\{ \tilde{\mathcal{X}}_{M}, \tilde{\mathcal{A}}_{M}, \tilde{\mathbf{w}}_{M} \}} \mathrm{GNN}(\{ \tilde{\mathcal{X}}_{M}, \tilde{\mathcal{A}}_{M}, \tilde{\mathbf{w}}_{M} \}; \Theta_{*}) $
     & Property \\ \midrule
     $\mathrm{HN\mbox{-}GFN^{\text{\cite{zhu2024sample}}}}$ & Search &
     $ \mathcal{S}_{i+1} = \mathcal{S}_{i} \cup \{ (M_{j}^{i}, f(M_{j}^{i})) \}_{j=1}^{b} $
     & Property \\ \midrule
     $\mathrm{DyMol^{\text{\cite{shin2024dynamic}}}}$ & Search &
     $\begin{cases}
         \mathbf{F}(\mathbf{z}_{x}, t) = \{ f_{1}(\mathbf{z}_{x}), f_{2}(\mathbf{z}_{x}), ..., f_{t}(\mathbf{z}_{x}) \} \\
         \mathbf{R}(x, t) = \sum_{i=1}^{t} w_{i}(t)f_{i}(\mathbf{z}_{x})
     \end{cases}$
     & Property \\ \midrule
     \end{tabular}
  \begin{minipage}{0.96\linewidth}
    \scriptsize 
    \vspace{3pt}
    \textbf{Notations}: 
    $\mathbf{z}_{\mathcal{T}}, \mathbf{z}_{\mathcal{G}}$ denote two-part latent representations of junction tree and graph structure. 
    $\mathcal{C}$ represents the generative conditions, $M_{\text{tag}_{i}}$ represents $i$-th atomic tags for the molecules' SMILES strings, $\mathrm{Trans}(\cdot)$ is the Transformer framework and $\text{Emb}$ is the standard input for the transformer, encompassing both word and positional embeddings. 
    $A_{k}$ is the reference arm. 
    $\odot$ denotes the element-wise product, "kn" and "ukn" are the abbreviations for "known" and "unknown", respectively. The noise $\epsilon \sim \mathcal{N}(0, I)$, $\tilde{w}$ is a conditional control strength parameter ($\tilde{w} \geq 0$), and $\tilde{w} = 0$ indicates unconditional generation. 
    $\mathcal{M}_{\theta}$ is the the molecular graph editing model and $\mathcal{S}$ is the constantly updated model training dataset. $p_{add}, p_{frag}, p_{del}$ are probability distributions with different operations on molecules. 
    $\mathrm{K}$ denotes the number of child nodes in the search procedure, $\mathcal{F}_{k}$ is the average reward of node $k$, $n_{k}$ is the number of times node $k$ is accessed, and $n$ is the iterative epochs. $\mathrm{d}$ is the reward vector dimension, $\lambda$ signifies an exploration parameter used to balance exploration and utilization, $\mathrm{P}(\cdot)$ denotes the set of nodes based on the Pareto optimality selection. 
    Meanwhile, $\mathbf{w}$ denotes the node weight vector and $\mathrm{GNN}(\cdot)$ is the graph neural network whose parameters are $\Theta$, 
    Finally, $\mathcal{S}$ is the constantly updated model training dataset, and $b$ represents the batch size. 
    $x$ is the element in the sequence of a molecule and $L$ is the maximum length of the sequence. 
  \end{minipage}
  \caption{Summary of molecular optimization methods.}
  \label{tab:MolOpt}
\end{table*}

\begin{table}[t!]
\centering
\renewcommand\arraystretch{0.95} 
\setlength\tabcolsep{3pt} 
\begin{threeparttable}
\resizebox{\columnwidth}{!}{
\begin{tabular}{cccr}
\toprule
\textbf{Datasets} & \# \textbf{Data Struc.}  & \# \textbf{Tasks} & \# \textbf{of Samples} \\
\midrule
\multicolumn{4}{c}{\textbf{Benchmark 1: TDC Datasets}$^{[1]}$} \vspace{0.1cm}  \\

BindingDB & Drug–Target & Reg. & 2,701,247 \\
DAVIS & Drug–Target &Reg. & 30,056 \\
KIBA & Drug–Target & Reg. & 118,254 \\
DrugBank & Drug–Drug &  Cla. & 191,808 \\
TWOSIDES & Drug–Drug & Cla. & 4,651,131 \\
GDSCv2 & Drug–Cell & Reg. & 243,466 \\
\midrule

\multicolumn{4}{c}{\textbf{Benchmark 2: TUDataset}$^{[2]}$} \vspace{0.1cm}  \\
QM9 & Drug(Qm., 3D) & Reg. & 133,885 \\
ZINC & Drug(Qm.) & Reg. & 249,456 \\
\midrule

\multicolumn{4}{c}{\textbf{Benchmark 3: Open Graph Benchmark}$^{[3]}$} \vspace{0.1cm}  \\
HIV & Drug(Bio.) & Cla. & 41,913 \\
Tox21 & Drug(Physio.)  & Cla. & 8,014 \\
ToxCast & Drug(Physio.)  & Cla. & 8,615 \\
BBBP & Drug(Physio.) & Cla. & 2,053 \\
PCQM4Mv2$^{[5]}$ & Drug(Qm.) &  Reg. & 3,746,619 \\
\midrule

\multicolumn{4}{c}{\textbf{Benchmark 4: MoleculeNet}$^{[4]}$} \vspace{0.1cm}  \\
PCBA & Drug(Bio.) & Cla. & 439,863 \\
MUV & Drug(Bio.) & Cla. & 93,127 \\
BACE & Drug(Bio.) & Cla. & 1,522 \\
SIDER & Drug(Physio.)  & Cla. & 1,427 \\
ClinTox & Drug(Physio.)  & Cla. & 1,491 \\
QM7 & Drug(Qm., 3D) & Reg. & 7,165 \\
QM8 & Drug(Qm., 3D) & Reg. & 21,786 \\
ESOL & Drug(Phys. Chem.) & Reg. & 1,128 \\
FreeSolv & Drug(Phys. Chem.) & Reg. & 643 \\
Lipophilicity & Drug(Phys. Chem.) & Reg. & 4,200 \\
\midrule

\multicolumn{4}{c}{\textbf{Others}} \\
MD17/22$^{[6]}$ & Drug(Qm.) &  Reg. & 99,999-627,983 / 5,032-85,109 \\
OC20/22$^{[7]}$ & Drug(Qm.) &  Reg. &  1,281,040 / 62,331 \\
Molecule3D$^{[8]}$ & Drug(Qm.) &  Reg. & 3,899,647 \\
3BPA$^{[9]}$ & Drug(Qm.) &  Reg. & 500 \\
ANI-1$^{[10]}$ & Drug(Qm.) &  Reg. & 24,416,306 \\

 &  &  & \multicolumn{1}{r}{\textbf{Drugs / Targets}} \\
LIT-PCBA$^{[11]}$ & Drug–Target &  Cla. & \multicolumn{1}{r}{415,225 / 15} \\
C.elegans$^{[12]}$ & Drug–Target &  Cla. & \multicolumn{1}{r}{1,434 / 2,504} \\
Human$^{[13]}$ & Drug–Target &  Cla. & \multicolumn{1}{r}{1,052 / 852} \\
DUD-E$^{[14]}$ & Drug–Target &  Cla. & \multicolumn{1}{r}{22,886 / 102} \\
Metz$^{[15]}$ & Drug–Target &   Reg. & \multicolumn{1}{r}{1,423 / 170} \\
Karimi$^{[16]}$ & Drug–Target &  Reg. & \multicolumn{1}{r}{3,672 / 1,287} \\
BioSNAP$^{[17]}$ & Drug–Target &  Reg. & \multicolumn{1}{r}{4,510 / 2,481} \\
 &  &  &  \multicolumn{1}{r}{\textbf{Drugs / Total}} \\
DRUGCOMBO$^{[18]}$ & Drug–Drug &  Cla. & \multicolumn{1}{r}{3,242 / 49,392} \\
KEGG$^{[19]}$ & Drug–Drug &  Cla. & \multicolumn{1}{r}{1,295 / 56,983} \\
SIDER$^{[20]}$ & Drug–Drug &  Cla. & \multicolumn{1}{r}{1,430 / 139,756} \\
ChCh-Miner$^{[21]}$ & Drug–Drug &  Cla. & \multicolumn{1}{r}{1,322 / 48,514} \\
OBG-biokg$^{[22]}$ & Drug–Drug &  Cla. & \multicolumn{1}{r}{808 / 111,520} \\
ZhangDDI$^{[23]}$ & Drug–Drug &  Reg. & \multicolumn{1}{r}{548 / 48,548} \\

 &  &  & \multicolumn{1}{r}{\textbf{Drugs / Cells}} \\
CCLE$^{[24]}$ & Drug–Cell & Reg. & \multicolumn{1}{r}{24 / 479} \\
NCI-60$^{[25]}$ & Drug–Cell & Reg. & \multicolumn{1}{r}{50,000 / 60}  \\

\bottomrule

\end{tabular}
}
\end{threeparttable}

\begin{minipage}{\linewidth} \scriptsize
\vspace{3pt}

$^{[1]}$\href{https://tdcommons.ai}{Therapeutics data commons dataset}; 
$^{[2]}$\href{https://chrsmrrs.github.io/datasets/}{The collection of benchmark datasets for learning with graphs};
$^{[3]}$\href{https://ogb.stanford.edu}{The open graph benchmark}; 
$^{[4]}$\href{https://moleculenet.org}{The benchmark for molecular machine learning}; 
$^{[5]}$\href{https://ogb.stanford.edu/docs/lsc/pcqm4mv2/}{Molecular dynamics benchmark for evaluating force fields}; $^{[6]}$\href{http://www.sgdml.org}{Two molecular dynamics datasets}; 
$^{[7]}$\href{https://fair-chem.github.io/index.html}{Two DFT-based molecular datasets for computational chemistry};
$^{[8]}$\href{https://github.com/divelab/MoleculeX}{A 3D geometries benchmark};
$^{[9]}$\href{https://service.tib.eu/ldmservice/dataset/3bpa-dataset}{A benchmark dataset for equivariant many-body interaction operations};
$^{[10]}$\href{https://materials.colabfit.org/id/DS_p4evspy1ntcs_0}{A dataset of 20 million conformations}; 
$^{[11]}$\href{https://drugdesign.unistra.fr/LIT-PCBA/}{A dataset with high-confidence data}; 
$^{[12]}$\href{https://snap.stanford.edu/data/C-elegans-frontal.html}{A Caenorhabditis elegans database};
$^{[13]}$\href{https://github.com/peizhenbai/DrugBAN/tree/main/datasets}{The human DTI dataset};
$^{[14]}$\href{https://dude.docking.org/}{A database of useful (docking) decoys-Enhanced};
$^{[15]}$\href{https://service.tib.eu/ldmservice/dataset/metz}{The G protein-coupled receptor focused dataset};
$^{[16]}$\href{https://drive.google.com/file/d/1_iZ8B1JZkCKmKlQNewOCr3kbnWfAIc-r/view}{A DTA and side effect dataset}; 
$^{[17]}$\href{https://snap.stanford.edu/biodata/datasets/10002/10002-ChG-Miner.html}{The biomedical network dataset}; 
$^{[18]}$\href{https://drive.google.com/file/d/1_iZ8B1JZkCKmKlQNewOCr3kbnWfAIc-r/view}{A drug combination synergy database}; 
$^{[19]}$\href{https://www.genome.jp/kegg/}{The gene\&genome database};
$^{[20]}$\href{http://sideeffects.embl.de/}{A drug-side effect association database};
$^{[21]}$\href{https://snap.stanford.edu/biodata/datasets/10001/10001-ChCh-Miner.html}{A DDI network dataset};
$^{[22]}$\href{https://ogb.stanford.edu/docs/linkprop/}{The biological knowledge with gene-drug-disease triples};
$^{[23]}$\href{https://github.com/zw9977129/drug-drug-interaction/tree/master/dataset}{A multi-source dataset};
$^{[24]}$\href{https://sites.broadinstitute.org/ccle}{The cancer cell line encyclopedia}; 
$^{[25]}$\href{https://dtp.cancer.gov/discovery_development/nci-60/}{The NCI-60 human tumor cell lines screen}.


\textbf{Notations}: Quantum mechanics (Qm.); Biophysics (Bio.); Physical chemistry (Phys. Chem.); Physiology (Physio.). The dataset statistics are accurate as of February 4, 2025.

\end{minipage} 
\caption{Common datasets for small molecule drug discovery.}
\label{Datasets}
\end{table}

\subsection{Molecular Generation}
\label{3.5}
The chemical space is vast, with over $10^{33}$ possible compound structures \cite{zhu2023molhf}, making traditional methods slow and labor-intensive. Computational MG methods have thus become essential to accelerate molecule discovery with desired properties.


Table \ref{tab:MolGen} provides a comprehensive summary. Initially, MG methods were largely unconditional random generation \cite{huang20223dlinker,jo2022score}, which failed to meet practical requirements. As a result, conditional MG methods have emerged as the mainstream approach. Key trends include 1) Property-conditioned generation \cite{zhu2023molhf}, which targets molecules with specific properties. 2) Molecular (sub)structure-conditioned generation, which uses predefined structural elements for design. 3) Target-conditioned generation \cite{guan2023decompdiff}, which focuses on molecules with high binding affinity to biological targets. 4) Phenotype-conditioned generation, which guides molecular design towards desired biological outcomes. Diffusion models have recently become dominant in MG \cite{lee2023exploring}, as they iteratively refine molecular structures from noise, offering more flexibility and integration with conditional frameworks for targeted molecular design.

Despite these advancements, poor interpretability remains a challenge. Most DL-based models operate as black boxes, making it difficult to understand how molecular structures are generated or which features influence specific properties. This limits their reliability and practical application in drug discovery, where understanding structure–property relationships is essential. Addressing this requires developing more interpretable frameworks that incorporate mechanistic insights and knowledge-driven constraints.

\subsection{Molecular Optimization}
\label{3.6}

MO task plays a key role in drug discovery and material design, refining molecular structures to meet specific functional, chemical, and biological criteria. Unlike MG methods, which focuses on creating new molecules, MO adjusts existing molecules to enhance their properties for particular applications, such as binding affinity, solubility, stability, and toxicity.


Table \ref{tab:MolOpt} provides a comprehensive summary. MO approaches can be broadly classified into two categories: generative models and search-based methods. Generative models like VAE, generative adversarial networks (GAN), flow-based methods, and diffusion models optimize molecules by manipulating their latent representations or refining structures iteratively from noise \cite{huang2024dual,xiong2024text}. In contrast, search-based methods, which use evolutionary algorithms \cite{fudifferentiable}, navigate the chemical space more directly. Additionally, auto-regressive models construct molecules sequentially, allowing precise control over the optimization process \cite{wu2024leveraging}. While early methods focused on latent space optimization, search-based approaches have gained popularity due to their superior interpretability and stability. Evolutionary algorithms evolve and eliminate molecules from a candidate set, while scoring functions modify molecules until the optimization objective is achieved. Other search-based methods map molecules to a high-dimensional solution space for optimization.


Despite these advances, MO still faces challenges with poor interpretability, as the connection between structural modifications and property enhancements remains unclear. Many optimization techniques propose molecular changes without explicitly explaining the rationale, hindering the understanding of how modifications improve specific attributes.

\section{Datasets}
\label{sec:4}

Small molecule drug discovery datasets form the foundation for research in this field. Typically, these datasets include information on the small molecules' characteristics such as their chemical structures, biological activities, pharmacokinetic properties, and toxicity, spanning multiple stages from target identification and lead compound screening to drug optimization. To facilitate research and application, these datasets are categorized based on their data structures and task types, with corresponding volumes provided. Since these datasets are suitable for molecular generation and optimization tasks, Table \ref{Datasets} does not specifically list the datasets related to these two tasks. Additionally, due to the large number and variety of datasets, we provide the access URLs for each dataset without citing the references individually.


We present several commonly used benchmark and non-benchmark datasets. As shown in Table \ref{Datasets}, the TDC datasets cover various tasks during small molecule drug discovery and are relatively mainstream. Additionally, QM9 and ZINC from TUDataset are molecular datasets that include a wide range of properties, such as quantum chemical properties, and are widely used in property prediction and conditional MG and MO tasks. Furthermore, the Open Graph Benchmark (OGB) and MoleculeNet databases include multiple molecular property datasets, which are commonly used for validating the performance of various molecular graph representation methods. For instance, MoleculeNet is a widely used benchmark dataset for molecular studies, encompassing four major categories: quantum mechanics, physical chemistry, biophysics, and physiology. Since the DRP, DDI, and DTI/DTA tasks investigate complex relationships between drugs and other entities (such as targets, cell lines, and drugs), the datasets involved are diverse and complex, and are thereby categorized under "Others."



\section{Challenges and Opportunities}
\label{sec:5}

\paragraph{Poor Interpretability.}

In the small molecule drug discovery field, achieving model interpretability through correlation (i.e., identifying which molecular features are strongly associated with the predicted outcomes) is relatively straightforward. This can be accomplished using traditional feature importance evaluation methods, such as SHapley Additive exPlanations or model-based attention mechanisms, which reveal the features that are strongly correlated with drug responses or drug–target interactions \cite{fudifferentiable,zhu2024sample}. For example, methods such as SubCDR \cite{subcdr} decomposed molecules into multiple subcomponents and quantified their contributions in the prediction, providing explanations and helping us understand the model's decision-making process. However, causal interpretability is more complex and challenging. Explaining the complex interactions between multiple entities requires understanding a molecule's influence on the activities and responses of another through specific mechanisms, necessitating more complex models and methods. Moreover, causal inference typically relies on multi-gene regulatory networks that involve interactions among various factors, requiring a deep understanding of biological background knowledge. Therefore, achieving causal interpretability may require more specific techniques and experimental methods. For instance, constructing gene regulatory networks, integrating interdisciplinary datasets, or combining these with wet-lab experimental validation could offer feasible solutions to address this challenge.

\paragraph{Low Out-of-Distribution Generalization Capability.}
\label{5.2}

The data combination patterns in drug discovery tasks, such as "drug+X" combinations, often give rise to potential OOD issues. Additionally, the MPP task commonly encounters OOD generalization challenges, particularly when datasets are divided based on molecular skeletons or certain similarity features, resulting in poor prediction performance for samples with significantly different characteristics. For "drug+X" combinations, the situation becomes more complex. Test scenarios can be categorized into four types based on whether the drugs and X in the test set have been observed in the training set. 1) Seen-Both: both drugs and X are present in the test and training sets. 2) Unseen-Drug: the drugs in the test set are not present in the training set. 3) Unseen-X: the X in the test set are not present in the training set. 4) Unseen-Both: both drugs and X in the test set are not present in the training set. All cases except Seen-Both represent OOD problems. 


To improve model generalizability in OOD scenarios, several strategies may be useful. First, data augmentation and generative models, such as VAEs and GANs, can be used to simulate various "drug+X" combinations, introducing the model to a broader range of potential scenarios during training and enhancing its adaptability to new environments. Second, transfer learning can leverage knowledge from tasks with known drugs or targets and apply it to new ones, assisting the model to address unseen drugs or X. Finally, domain adaptation techniques have recently demonstrated the ability to reduce distributional discrepancies between training and testing datasets \cite{MSDA,code-ae} by learning the mapping relationships between the source and target domains, further enhancing the model's predictive ability on new data.

\paragraph{Model Training and Laboratory Validation Gap.}


Current methods typically make predictions directly after training, and those used in real-world applications often lack the ability to continuously learn and adapt to new experimental results. This limitation restricts the model's ability to effectively utilize new experimental data. Specifically, there is a noticeable gap in performance between preclinical datasets and actual laboratory drug development processes. This is demonstrated by issues such as feature drift and sample bias. To address this, online and incremental learning techniques can be incorporated, allowing models to be updated continuously during real-world applications. For example, online reinforcement learning, commonly used in DL systems like large language models (LLMs), can adaptively adjust model parameters in response to new data streams, enhancing the model's ability to generalize and adapt to dynamic environments and new data.



\paragraph{Lack of Fair Benchmarking Results.}

Although numerous machine learning methods have been developed, a lack of standardized and fair benchmarking results for key tasks exists. For example, many studies only use a small portion of the available data for training, limiting the model's scalability. Therefore, establishing standardized computational evaluation protocols and large-scale benchmark testing is critical for advancing drug discovery. By developing large-scale multi-task and multi-modal datasets and conducting standardized benchmarking tests, fair comparisons between different models and methods can be promoted, ultimately enhancing model scalability and generalizability.

\bibliographystyle{ieeetr_ICWS}  
\bibliography{IEEEtran}

\end{document}